\documentclass[11pt,a4paper]{article}
\usepackage{times,latexsym}
\usepackage{url}
\usepackage[T1]{fontenc}

\usepackage[acceptedWithA]{tacl2021v1}
\usepackage{xspace,mfirstuc,tabulary}

\usepackage{amsmath}

\newif\iftaclinstructions
\taclinstructionsfalse 
\iftaclinstructions
\renewcommand{\confidential}{}
\renewcommand{\anonsubtext}{(No author info supplied here, for consistency with
TACL-submission anonymization requirements)}
\newcommand{\instr}
\fi

\iftaclpubformat 

\else

\fi

\usepackage{makecell}
\usepackage{xcolor,colortbl}
\usepackage{color,soul}
\usepackage{graphicx,multirow}
\usepackage{booktabs}
\usepackage{comment}
\usepackage{dsfont}
\usepackage{amsmath}
\usepackage{amsfonts}
\usepackage{amssymb}
\usepackage{mathtools}
\usepackage{subcaption}
\usepackage{fixltx2e}
\DeclareRobustCommand{\hlyellow}[1]{{\sethlcolor{yellow!60}\hl{#1}}}
\DeclareRobustCommand{\hlgray}[1]{{\sethlcolor{gray!20}\hl{#1}}}

\newcommand\ie{\emph{i.e.}}
\newcommand\eg{\emph{e.g.}}
\newcommand\name{\textsc{MEG}}

\title{MEG: Medical Knowledge-Augmented Large Language Models\\for Question Answering}

\author{
}
  
\author{
  Laura Cabello$^\diamond$,
  Carmen Martin-Turrero$^\dagger$, 
  Uchenna Akujuobi$^\dagger$, \\
  \textbf{Anders Søgaard}$^\diamond$ \and
  \textbf{Carlos Bobed}$^\circ$
  \\
  $^\diamond$University of Copenhagen, Denmark \\
  $^\dagger$Sony AI, Barcelona, Spain \\
  $^\circ$University of Zaragoza, Spain
  \\
  \texttt{\{lcp, soegaard\}@di.ku.dk},
  \texttt{cmartur@gmail.com}, \\ \texttt{uchenna.akujuobi@sony.com},
  \texttt{cbobed@unizar.es}
}

\date{}

\begin{document}
\maketitle

\begin{abstract}
Question answering is a natural language understanding task that involves reasoning over both explicit context, and unstated relevant domain knowledge. 
Despite the high cost of training, large language models (LLMs)---the backbone of most modern question-answering systems---still struggle to reliably capture the nuanced relationships between concepts that are crucial for reasoning in specialized fields like medicine.
In this work, we present \name{}, a parameter-efficient approach for medical knowledge-augmented LLMs. 
\name{} uses a lightweight mapping network to incorporate knowledge graph embeddings into the LLM, enabling it to leverage external knowledge in a cost-effective way.
We evaluate our method on four popular medical multiple-choice datasets and show that LLMs \emph{i)} can effectively interpret knowledge graph embeddings and \emph{ii)} gain significant advantages from the factual grounding these embeddings provide.
\name{} attains an average of +6.7\% 
and +9.9\% accuracy over specialized models like BioMistral-7B and MediTron-7B, respectively. 
Finally, we show that \name{}'s performance remains robust to the choice of graph encoder.
\end{abstract}

\section{Introduction}
Large language models (LLMs) induce knowledge from vast text corpora.
Through self-supervised learning, these models capture deeply 
contextualized representations of input tokens that enable them to generalize to new tasks with remarkable performance.  
This, as well as their ability to write long coherent passages, has made LLMs incredibly popular, despite their considerable inference costs \citep{cheng-etal-2023-batch} and their concerning carbon footprint \citep{strubell-etal-2019-energy}.
Moreover, current LLMs face significant challenges with handling complex reasoning and ensuring trustworthiness \cite{Liu2023TrustworthyLA,pmlr-v235-huang24x} and factual consistency \cite{maynez-etal-2020-faithfulness,10.1145/3544548.3581318,tam-etal-2023-evaluating,hager2024evaluation}, essential to critical fields like healthcare.
While LLMs are poised to revolutionize our medical system, already performing well on medical licensing exams \cite{jin2020medqa,pmlr-v174-pal22a,singhal2023large,brin2023comparing} and other tasks \cite{NAZARIOJOHNSON20231004,van-veen-etal-2023-radadapt,tu2023generalistbiomedicalai,carl2024large}, there is still much room for improvement.

To improve reliability and reduce computational costs, 
researchers have experimented with training from mixtures of corpora and knowledge bases \citep{pan_et_alTGDK,10387715}. 
Knowledge Graphs (KGs), such as the Unified Medical Language System (UMLS) \cite{bodenreider2004UMLS}, are structured knowledge bases that explicitly store rich factual knowledge. KGs are good at capturing the nuances of complex data and can provide complementary information to LLMs, especially useful for tasks requiring structured understanding. 
However, this complementary information is introduced as a new modality: graphs. Therefore, it extends beyond the plain text that LLMs can naturally interpret, requiring additional mechanisms to process and integrate it effectively.
The potential of knowledge-augmented LLMs\footnote{In this work, we define a knowledge-augmented LLM as an LLM enhanced with KG embeddings (KGEs). KGEs are dense vector representations of graph entities \cite{JU2024106207survey}. Therefore, we also refer to knowledge-augmented LLMs as KGE-augmented LLMs throughout the paper.} outlines an interesting research paradigm that can alleviate current challenges of LLMs, and reduce the need of training ever-larger models \cite{hooker2024limitationscomputethresholdsgovernance}. However, how to effectively model interactions between LLMs and KGs remains an open question.

Recent efforts have focused on self-supervised methods for jointly training
graph neural networks and pretrained language models \cite{yang2021graphformers,chien2022node,brannon-etal-2024-congrat}. Others \cite{DRAGON_NEURIPS2022,tang2023graphgpt,plenz-frank-2024-graph}, propose new model architectures to leverage the two modalities, graph and text, during pretraining. These methods learn deep interactions over text and graph, but they require carefully curated pretraining data, are mainly studied for graph-oriented tasks \cite{yang2021graphformers,chien2022node,tang2023graphgpt}, or are yet to be adapted to a generative framework \cite{DRAGON_NEURIPS2022,plenz-frank-2024-graph}.

In this work, we introduce \name{}, a parameter-efficient approach 
to MEdical knowledGe-augmented LLMs for question answering (QA).
MEG eliminates the need to train base LLMs for specialized domains by incorporating relevant KGEs at inference time. 
To achieve this, we design a lightweight mapping network that unidirectionally translates KGEs into the LLM's vector space, allowing the model to interpret these embeddings, which, in turn, further conditions its response generation. Notably, KGEs are injected directly at token level, after the model’s embedding layer.
We experiment with Mistral-Instruct (7B) \cite{jiang2023mistral7b} and Llama-3-Instruct (8B) \cite{dubey2024llama3herdmodels} as our base models, and report results with our best setup for KGEs: a KG encoder based on GraphSAGE \cite{NIPS2017_graphsage} combined with a simple Multilayer Perceptron (MLP) as mapping network.

In sum, our \textbf{contributions} are as follows:\footnote{Code and KGEs will be published upon acceptance.} 
\emph{i)} We introduce \name{}, a novel approach to \textbf{knowledge-augmented LLMs based on KGEs}\footnote{While we frame our approach in the medical domain, we would like to remark that it is domain independent given that it is built on any large KG. However, as we focused in this domain given the abundance of previous results and datasets, we refrain from making further generalized claims (in spite of there being strong evidences supporting them).}. MEG is an efficient approach to domain specialization of models, as it updates only a small fraction of the LLM's parameters while providing access to a large, reliable knowledge source (a KG). 
\emph{ii)} We conduct extensive evaluation on the four popular multiple-choice QA datasets from the MultiMedQA \cite{singhal2023large} clinical benchmark, and \textbf{demonstrate the effectiveness of integrating pretrained KGEs into LLMs} for medical question answering. Specifically, \name{} surpasses well-establish biomedical LLM baselines like BioMistral-7B \cite{labrak-etal-2024-biomistral} or MediTron-7B \cite{chen2023meditron70bscalingmedicalpretraining}, which have followed a costly continued pretraining of the base LLMs on curated biomedical data.
\emph{iii)} We \textbf{provide insights into \name{}'s dual modality} by examining the contributions of each module and comparing embedding spaces of KGEs and LLMs. We intuitively explain the shifts in the LLM's representations that drive \name{}'s stronger performance.

\section{Related Work}

\paragraph{Medical Language Models}
Current state-of-the-art (SOTA) in medical QA benchmarks like MedQA \cite{jin2020medqa}, PubMedQA \cite{jin-etal-2019-pubmedqa} or MedMCQA \cite{pmlr-v174-pal22a} belongs to close-sourced models of unknown size like Med-Gemini \cite{saab2024capabilitiesgeminimodelsmedicine}, Med-PaLM2 \cite{singhal2023expertlevelmedicalquestionanswering} or GPT-4 \cite{nori2023generalistfoundationmodelsoutcompete}.
Popular open-source LLMs in biomedicine include MedAlpaca \cite{han2023medalpacaopensourcecollection} and PMC-LLaMA \cite{wu2023pmcllamabuildingopensourcelanguage} based on Llama \cite{touvron2023llamaopenefficientfoundation}, MediTron \cite{chen2023meditron70bscalingmedicalpretraining} based on Llama-2 \cite{touvron2023llama2openfoundation}, or BioMistral \cite{labrak-etal-2024-biomistral} based on Mistral-Instruct \cite{jiang2023mistral7b}. These models continue pretraining the base general-purpose models on curated medical corpora.
More recently, 
\citet{kim2024smalllanguagemodelslearn} present the Meerkat models 
trained with chain-of-thought \cite{10.5555/3600270.3602070} synthetic data. Meerkat-7B outperforms the previous best 7B models across several medical benchmarks. However, it takes eight 80G A100 GPUs and 1.5 days to complete training.
In contrast, our approach 
is among the first to leverage pretrained medical KGEs to condition the LLM's generation and can be trained on four A10G GPUs within a few hours (see \S~\ref{sec:experiments} for details).

\paragraph{Knowledge-Augmented Language Models}
Bringing together LLMs and KGs is an active line of research that has gained increasing attention from both academia and industry \citep{pan_et_alTGDK,10387715}. Among numerous efforts in this area, \citet{zhang-etal-2019-ernie,DRAGON_NEURIPS2022,tang2023graphgpt,ZHU2023119369}, to name a few, propose different methods for combining text and graphs during pretraining. 
Parallel to these lines of work, \citet{sarmah2024hybridragintegratingknowledgegraphs,edge2024localglobalgraphrag,hu2024graggraphretrievalaugmentedgeneration,mavromatis2024gnnraggraphneuralretrieval} approach the integration of LLMs and KGs through retrieval-augmented generation (RAG) \cite{NEURIPS2020_6b493230}. 
However, the deployment of such knowledge-augmented LLMs for medical QA remains understudied.
Our work fills this gap and presents a novel approach to medical knowledge-augmented LLMs based on KGEs.
We note that \name{} may resemble a sort of RAG system, where an LLM leverages knowledge from an external database of KGEs. 
However, our approach does not retrieve document chunks to include them in the textual prompt. Instead, MEG fetches KGEs of named entities from the user's prompt and injects these embeddings into the LLM, enriching the query's context and guiding answer generation. 
For a comprehensive benchmark of RAG in medical QA, we refer the reader to \citet{xiong-etal-2024-benchmarking}.

\section{Problem Formulation}\label{sec:problem}
Our objective is to augment a general purpose LLM with specialized knowledge, so it can generate accurate answers without the need of fine-tuning to the new domain. We propose to use a large KG as the source of specialized knowledge. The reason for this is two-folded: a KG is a reliable source of factual knowledge and it captures structured relationship between concepts, that can potentially augment the LLM's representation of such concepts. This design choice marks the multimodality of our solution: \emph{text} generation will be conditioned on embedded representations of \emph{graph} concepts. 

Thus, our main task is to augment an LLM with KG embeddings to answer 
questions in a relevant domain. 
Our proposed approach, \name{}, consists of four key components:\vspace{-2mm}

\begin{enumerate}
    \item[\emph{i)}] A \textbf{KG encoder} to represent knowledge graph entities in a continuous vector space, while preserving their semantic meaning.\vspace{-2mm}
    \item[\emph{ii)}] An \textbf{instruction-tuned language decoder} capable of generating textual answers.\vspace{-2mm}
    \item[\emph{iii)}] A \textbf{mapping function $f_{\text{k}}$} that transforms the output of the KG encoder into a representation that can be used by the language decoder. $f_{\text{k}}$ is parameterized by a neural network. Thus, we interchangeably use the term mapping network.\vspace{-2mm}
    \item[\emph{iv)}] A \textbf{KG grounding module} that detects textual entities and grounds them in graph entities.
\end{enumerate}

\begin{figure}[t]
    \centering
    \includegraphics[width=\columnwidth]{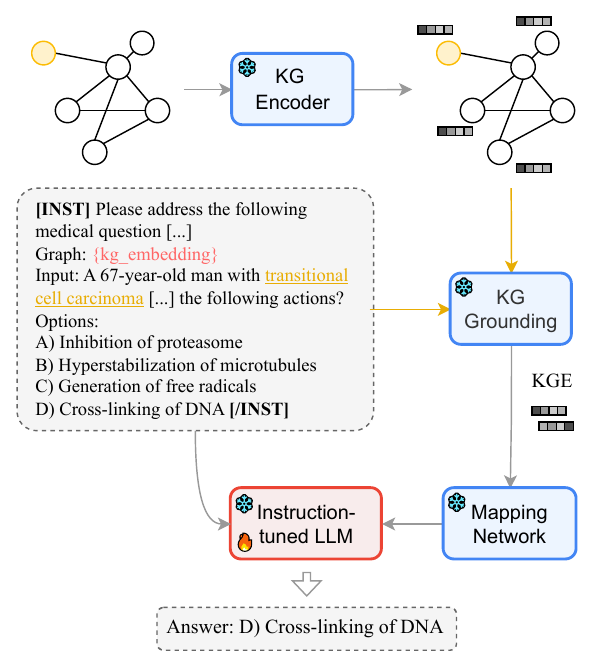}
    \caption{\name{} leverages a pretrained KG encoder and an LLM. During an initial phase of training, \name{} learns a mapping network to convert relevant graph features (KGEs) retrieved by the grounding module into token embeddings. During downstream fine-tuning, only a subset of the LLM's weights is updated, while the embedding layer and mapping network remain frozen. At inference, the LLM takes the text and the mapped KGEs as input and generates a response.}
    \label{fig:main}
    \vskip -0.1in
\end{figure}

Figure~\ref{fig:main} depicts our full pipeline. We carefully investigate the design of these components and how they interact with each other (\S~\ref{sec:model}). 
The mapping network is jointly trained with the LLM to learn an appropriate transformation of KGE, and it is frozen in downstream applications (\S~\ref{sec:training}).

\paragraph{Definitions.} 
A generic dataset for multiple-choice question answering (QA) consists of examples with a context paragraph, a question and a candidate answer set, all expressed in text. Given a QA example, each prompt $W$ is the concatenation of context, question and candidate answer set. We denote the sequence of tokens (words) in \(W\) as \( \{w_1, \ldots, w_S\} \), where $S$ is the maximum  sequence length. We denote the sequence of tokens (vectors) in the language model embedding space as \( W_e = \{w_{e1}, \ldots, w_{eS}\} \).

We define a knowledge graph (KG) as a directed graph $G=(V, E)$, where $V$ is the set of entity nodes, and $E \subseteq V \times R \times V$ is the set of edges (triples) that connect nodes in $V$, with $R$ being the set of relation types.
Each triple 
$(s, p, o)$ in a KG represents a knowledge fact, 
such as $($\textsc{Headache}, \textsc{is\_a}, \textsc{Cephalgia}$)$.
A KGE $e$ is a mathematical representation that maps each entity \( v \in V \) 
and each relation \( r \in R \) of a directed knowledge graph $G$ to low-dimensional vectors in $\mathbb{R}^g$, preserving the semantic relationships within the graph.

Finally, we define a KGE-augmented language model to be a function
\(f_{\text{l}}(W_e \oplus f_{\text{k}}(X))\)\footnote{Formally, its domain is the set of sequences of elements $x_i \in \mathbb{R}^l$.} with \(f_{\text{l}} \in \mathbb{R}^l\),
where $f_{\text{k}}(X)$ is a set of KGEs, $\{e_{1}, \ldots, e_{N}\}$ with \(e_i \in \mathbb{R}^g \), that has been mapped to the LLM's space using a learned mapping function $f_k:\mathbb{R}^g \rightarrow \mathbb{R}^l$. The language model $f_{\text{l}}$ concatenates these representations to the token word embeddings $W_e$ to perform downstream tasks in the fine-tuning steps.
A language model is a special case of a KGE-augmented language model with no KGE (N=0).


\section{Method}\label{sec:method}
\subsection{\name{}}\label{sec:model}
\name{} combines a pretrained KG encoder and a pretrained LLM by means of an intermediate mapping network (see Figure~\ref{fig:main}). The KG encoder, which is trained separately on a large medical KG\footnote{Specifically, we use \textbf{UMLS} \cite{bodenreider2004UMLS}, a widely-used KG in biomedicine with $\sim$300K nodes (entities) and one million edges in total.}, provides graph embeddings that are directly fed to the mapping network. 
The mapping network outputs embeddings that are injected into the LLM after its embedding layer, influencing the LLM's answer generation.


\paragraph{Knowledge Graph Encoder}\label{sub:kgencoder}
The KG encoder is trained up-front over the selected graph to generate KGEs. We choose GraphSAGE \cite{NIPS2017_graphsage} as our preferred KG encoder. In \S~\ref{sec:ablation} we present an ablation study with random-walk-based, energy-based translational, and message-passing encoders.

\begin{figure}[h]
\begin{center}
\includegraphics[width=\columnwidth]{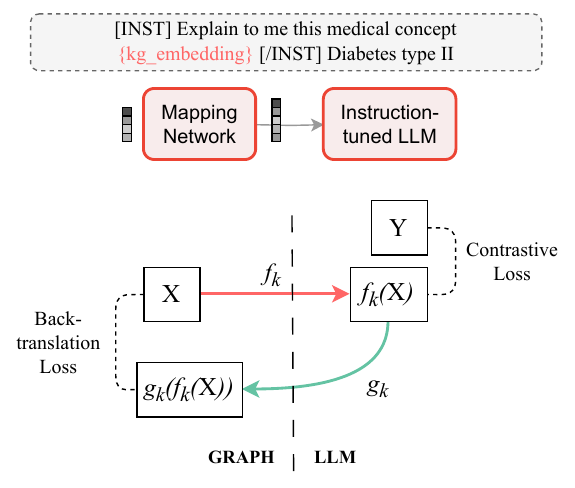}
\end{center}
\caption{
$f_{\text{k}}$ and $g_{\text{k}}$ are embedding transfer functions. $f_{\text{k}}$ takes a set of KGEs $X$ (\ie, node entities) as input, and outputs a mapping of $X$ to the LLM's vector space. $Y$ is the set of averaged token embeddings of entities in the LLM space. During training, $g_{\text{k}}$ prevents degenerated transformation of graph embeddings. The dashed lines indicate the input for the objective losses.
}
\label{fig:mapping}
\end{figure}

\paragraph{Mapping Network} 
The mapping function $f_{\text{k}}$ transforms a sequence of graph features from the KG encoder into a sequence that can be consumed by the LLM. We parameterize $f_{\text{k}}$ as an MLP with four hidden layers of size $d_h=128$. 
In particular, a set of graph embeddings is transformed from $d_g=256$ to $d_{l}=4096$ after a series of non-linear transformations through the hidden layers of $f_{\text{k}}$. We denote the initial embedding sets as \( X=\{x_i\}^N_{i=1} \), \( Y=\{y_j\}^{N}_{j=1} \), \(x_i \in \mathbb{R}^{d_g}, y_j \in \mathbb{R}^{d_l}\), being $x_i$ the KGEs, $y_j$ the averaged token embeddings of the entity in the LLM, and $N$ the total number of graph embeddings (entities).
We further denote the set of mapped embeddings as \( f_{\text{k}}(X) := \{f_{\text{k}}(x_i)\}^n_{i=1} \).

The goal is to learn the mapping $f_{\text{k}}$ that transforms $X$ to the LLM's vector space, while preserving its semantic meaning and structural information. Rather than minimizing the sum of squared differences between $f_{\text{k}}(X)$ and $Y$, we aim at positioning each $x_i$ in the neighborhood of its counterpart in $Y$. Pursuing an exact match of space distributions, such as through a Procrustes transformation \cite{schonemann1966procrustes,Gower1975}, would disregard the structural knowledge encoded in $X$.

To achieve this, we design an architecture similar to \citet{xu-etal-2018-unsupervised-cross} with two mappings \( f_{\text{k}}: X \rightarrow Y \) and \( g_{\text{k}}: Y \rightarrow X \), as illustrated in Figure~\ref{fig:mapping}.
We construct an instruction dataset with labels from UMLS's entities to teach the LLM to interpret the transformed graph embedding \( f_{\text{k}}(x_i) \). 
Figure~\ref{fig:mapping} (top) shows an example of an instruction, where the placeholder \texttt{\{kg\_embedding\}} is replaced by \( f_{\text{k}}(x_i) \) at token level. 
We train the full network jointly with the LLM\footnote{We conducted experiments by training the mapping network and LLM separately on UMLS. However, this approach resulted in worse performance on the downstream tasks tested.}.
Our loss function consists of three parts: a standard next-token prediction objective (\textbf{cross-entropy loss} \(\mathcal{L_\text{ce}}\)), and a sum of a contrastive loss and a back-translation loss to optimize the mapping network. Specifically,\vspace{-2mm}

\begin{itemize}
    \item 
    Given a batch \(X_b\) including a positive pair of examples $x_i$ and $x_j$, a contrastive objective \cite{1640964} is a function whose value is low when $x_i$ is similar to $x_j$ and dissimilar to all others, which are considered negative pairs for $x_i$. We employ a popular contrastive self-supervised learning objective \cite{NIPS2016_6b180037,oord2019representationlearningcontrastivepredictive,He2019MomentumCF}, dubbed as \textbf{NT-Xent loss} by \citet{chen2020simpleframeworkcontrastivelearning}. 
    NT-Xent uses dot product as similarity measure, and computes a normalized temperature-scaled cross-entropy loss for a positive pair as follows, 
    \begin{equation}
        \ell_{i,j} = -\log \frac{\exp(x_i \cdot x_j / \tau)}{\sum_{k=1,k \neq i}^{B} \exp(x_i \cdot x_k / \tau)},
    \end{equation}
    where B is the batch size and $\tau$ is the temperature. We set the hyper-parameter~\(\tau~=~1.0\)~\footnote{We also evaluate \(\tau=0.5\) as in \citet{chen2020simpleframeworkcontrastivelearning} and \(\tau=0.07\) as in \citet{He2019MomentumCF}. We choose the final value of \(\tau=1.0\) based on accuracy attained on a zero-shot setting on the validation split in MedQA.}. The final loss \(\mathcal{L_\text{c}}\) is computed across all positive pairs in a batch, summed across all batches. 
    Intuitively, the contrastive loss serves as an unsupervised objective function for training the  network to bring similar entities closer together in $Y$ and push dissimilar ones apart.
    
    \item We also employ a \textbf{back-translation loss} for preventing degenerated transformation. We enforce that the graph embedding after the forward and the backward transformation should not diverge much from its original direction. Following \citet{xu-etal-2018-unsupervised-cross}, we choose the back-translation loss based on cosine similarity. Note that our primary goal is to optimize the forward mapping \( f_{\text{k}}: X \rightarrow Y \). Thus, we do not control for back-translation in the reversed path, \( g_{\text{k}}: Y \rightarrow X \),
    \begingroup
    \small
    \begin{equation}
        \mathcal{L_\text{bt}}(f_{\text{k}}, g_{\text{k}}) = \sum_{i} \left( 1 - \cos(x_i, g_{\text{k}}( f_{\text{k}}(x_i) ) ) \right)
    \end{equation}
    \endgroup
\end{itemize}

Thereby, when training the mapping network jointly with the LLM, we minimize the following objective function: 
\begin{equation}\label{eq:loss}
    \mathcal{L} = \alpha \mathcal{L_\text{c}} + \beta \mathcal{L_\text{bt}} + \mathcal{L_\text{ce}},
\end{equation}

where $\alpha$ and $\beta$ are scalar hyperparameters to weight each objective in the transformation process. Our network design achieves a good trade-off between expressivity and parameter count, totaling 1.22M parameters. After the mapping is learnt, we freeze the network's weights and the LLM's embedding layer during the fine-tuning to downstream tasks (Phase II of training). The backwards transfer network $g_{\text{k}}$ is disconnected and only $f_{\text{k}}$ is used to do the mapping.

\paragraph{Grounding Module}
The grounding module takes textual data $W$ as input and links mentioned entities in $W$ to their corresponding nodes in the knowledge graph $G$, generating the corresponding entity mentions' KGEs for the LLM to use. These KG embeddings will enrich the question's context, helping to guide the answer. A critical step in this process is medical entity disambiguation \cite{10.1145/3448016.3457328,lu-etal-2024-medical}, which involves detecting named entities in $W$ and linking them to their unique counterparts in $G$. Since an entity can be referred to in multiple ways, for instance `heart attack' and `myocardial infarction', this step standardizes variations 
by linking them to a unique identifier in $G$ 
(following with the previous example, both mentions correspond to Concept Unique Identifier (CUI) `C0155626' in UMLS). 
This grounding ensures retrieval of relevant information for each example.
We use the entity linker presented in \citet{neumann-etal-2019-scispacy}\footnote{We use the last version of scispaCy (v2.5.0), which supports linking to UMLS and has near 3M unique concepts.}, which covers 99\% of the concepts mentioned in the MedMentions dataset \cite{mohan2019medmentionslargebiomedicalcorpus} and 86\% of the concepts mentioned in the MedQA dataset \cite{jin2020medqa}. These two datasets provide ground truth UMLS annotations.

\subsection{Training}\label{sec:training}
We aim to train \name{} to achieve competent results on medical question answering benchmarks while minimizing computational cost. To do this, 
We perform an initial training phase for embedding transfer learning.

\paragraph{Phase I: Embedding Transfer Learning}
We first learn the optimal transformation $f_{\text{k}}: X \rightarrow Y$ so that the mapped KG embeddings retain relevant information from the KG and can be effectively used by the LLM. As explained in \S~\ref{sec:model}, we create an instruction dataset from UMLS labels to guide the LLM in learning the relationship between its original representation of medical entities and their mapped graph embeddings. The train set contains 297,927 examples, following the same template shown in Figure~\ref{fig:mapping} for every entity label in UMLS.\footnote{We investigate whether data augmentation at this stage could lead to more accurate results in downstream tasks. We augment the initial $\sim$300K examples by creating new instructions with multiple entities, \eg `Explain to me these medical concepts: [\ldots]', to better match the setting from downstream tasks, which often include several entities per sample. This process doubles the dataset size, with an augmentation that normally distributes the number of entities per instruction between 2 and 10. Results are within \( \pm0.2 \) accuracy in MedQA compared to training without the augmented data. Due to the extra computational costs and minor (if any) gains, we did not explore this option further.} We train for one epoch jointly the mapping network and the LLM to minimize the objective from Eq.~\ref{eq:loss}.
At this stage, \name{} can leverage concepts from the KG to augment its domain knowledge and generate better, informed answers.

\paragraph{Fine-tuning on Downstream Tasks}
Given a medical multiple choice QA dataset, we fine-tune \name{} to answer the input question based on the textual content and information leveraged from the mapped KG embeddings. 
We format the input prompts $W$ as follows. For each example in the dataset, we concatenate the context (if any), question and candidate answer set following the pseudo-code shown in Figure~\ref{fig:instruction_template}.

At the token level, the placeholder embedding \texttt{\{kg\_embedding\}} is replaced with $N$ transformed KGEs 
produced by the mapping network\footnote{
In both training phases, we investigate the effect of injecting the mapped KGEs at the last layer of the LLM instead of after the embedding layer. These early experiments revealed little to no impact on zero-shot downstream accuracy, but slightly worsen the fine-tuned accuracy as measured on the validation set of MedQA with three random initialization seeds. This finding suggests that the LLM benefits from attending the external KGEs during fine-tuning, enabling more contextualized representations of these embeddings.}.
The mapping network weights remain frozen, and the backward network $g_{\text{k}}$ is effectively disabled, which previously served to regulate the learning of $f_{\text{k}}$ and prevent degenerated transformations. Similarly, the LLM's embedding layer is also frozen.
This approach reduces the computational complexity and speeds up fine-tuning.

\begin{figure}[h!]
    \centering
    \fbox{
        \begin{minipage}{0.9\columnwidth} 
        [INST] Please address the following medical question based on the Input text and any useful information you may find in the given concepts from a medical graph. \\
        \textbf{Input:} {{context}} {{question}} \\
        \textbf{Options:} \\
         \{\% for option in options \%\}  \\
         \{\{letter\}\}) \{\{text\}\} \\
         \{\% endfor \%\} \\
        Answer with the best option directly. Ignore irrelevant information. \\
        \textbf{Graph:} \{\{kg\_embeddings\}\} [/INST] \\
        \textbf{Answer:} \{\{correct\_option\}\} \\
        \end{minipage}
    }
    \caption{Template used to generate instructions for all QA datasets. The context is optional, depending on the dataset. At inference time, the text after [/INST] is generated by the language model. }
    \label{fig:instruction_template}
\end{figure}

\section{Experimental Details}\label{sec:experiments}
\paragraph{Data}
Following previous research on medical LLMs,
we evaluate \name{} on four well-known medical benchmarks that require extensive background knowledge. 
The first one is MedQA-USMLE (\textbf{MedQA}) \cite{jin2020medqa}, which consists of 10,178 train questions and 1,273 test questions, formatted with four choices each. The content was originally curated by experts from the US Medical License Exam. 
The second benchmark, \textbf{PubMedQA} \cite{jin-etal-2019-pubmedqa}, was collected from PubMed abstracts and includes 1,000 expert labeled question-answer pairs. The task is to produce a yes/no/maybe answer based on the question and an abstract as context. As previously done by others \cite{singhal2023large,chen2023meditron70bscalingmedicalpretraining,labrak-etal-2024-biomistral}, we use 500 random\footnote{We split the data following a similar distribution of answers between train and test splits.} samples for evaluation. The remaining 500 samples, though limited in size, serve as our only source of training data. 
We exclude the 211k artificially labeled yes/no samples provided by \citet{jin-etal-2019-pubmedqa} to avoid bias towards these two options. 
The third benchmark, \textbf{MedMCQA} \cite{pmlr-v174-pal22a}, contains 179,722\footnote{We detect 3,100 duplicate questions in the train split, which we remove.} train questions from Indian medical entrance exams. Due to the unavailability of answer keys for the test set, we follow others \cite{wu2023pmcllamabuildingopensourcelanguage,tu2023generalistbiomedicalai,labrak-etal-2024-biomistral} and report results on the validation set (4,183 questions).
Lastly, \textbf{MMLU-Medical} \cite{singhal2023large} includes 1,089 questions, each with four options, across six medical and biology-related categories drawn from \citet{hendrycks2021measuring}. 
Since this dataset only provides test data, we evaluate the generalization performance of \name{} fine-tuned on MedMCQA as in \cite{chen2023meditron70bscalingmedicalpretraining}. Thus, results on MMLU-Medical report out-of-distribution inference.

\paragraph{Training Details}
We initialize the KG node embeddings with token embeddings from SapBERT \cite{Liu2020SelfAlignmentPF}. SapBERT leverages contextualized embeddings from a pretrained BERT-based language model for biomedical KGs like UMLS. This initialization leads to improved performance compared to random embedding initialization. We train GraphSAGE with same hyperparameters as in \citet{NIPS2017_graphsage}.

During \textbf{phase I} of training, described in \S~\ref{sec:training},
we randomly initialize the mapping network and load the pretrained weights of the LLM. We fully train the mapping network and perform low-rank adaptation (LoRA, \citet{hu2022lora}) fine-tuning on every linear layer of the LLM (except for the embedding layer), while the remaining parameters are frozen.
This parameter-efficient tuning approach allows to learn the equivalent of 2\% of the model's parameters. Our full architecture results on 216M trainable parameters.
After training, we merge the LLM's updated parameters with the base model. 
Training takes 4h on 4 NVIDIA A10G GPUS using DeepSpeed\footnote{\url{https://www.deepspeed.ai/}} for distributed training. See Appendix~\ref{app:training_details} for further details.

After phase I, the models are evaluated in a zero-shot setup on downstream tasks, already able to use information from graph embeddings. 
To achieve optimal performance in downstream tasks, we fine-tune the models as in \S~\ref{sec:training}.
To allow batching, the number of KGEs injected to the LLM is fixed across samples\footnote{
The average number of ground entities per instance varies across datasets according to the median number of ground entities.
We set $N=20$ in MedQA, PubMedQA and `professional medicine' in MMLU-Medical; $N=3$ in MedMCQA and $N=2$ in the remaining categories from MMLU-Medical.}. If the grounding module retrieves more KGEs, we randomly select $N$. Otherwise, we add zero-padding.  
See Appendix~\ref{app:training_details} for training details.

\begin{table}[h!]
    \begin{center}
    \begin{tabular}{llc}
        & Model & Acc  \\ \toprule
        \multirow{3}{*}{ZS}
        &Mistral-Instruct-v0.1$^\dagger$ &42.3$_{\pm{0.3}}$\\ 
        &BioMistral$^\dagger$ &44.4$_{\pm{0.2}}$ \\
        &Mistral-Instruct-v0.1 w/ graph &40.4$_{\pm{0.4}}$ \\\midrule
        \multirow{3}{*}{FT}
        &Mistral-Instruct-v0.1$^\dagger$ &42.0$_{\pm{0.2}}$ \\
        &BioMistral$^\dagger$ &50.6$_{\pm{0.3}}$\\
        &Mistral-Instruct-v0.1 w/ graph &52.7$_{\pm{0.2}}$ \\\toprule
    \end{tabular}
    \end{center}
    \caption{Ablation study on the utility of the information encoded in knowledge graph triples expressed in natural language. We report accuracy on MedQA. ZS stands for `zero-shot'; FT stands for `fine-tuning'. $^\dagger$Results from \citet{labrak-etal-2024-biomistral}. }\label{tab:text_graph}
\end{table}

\section{Results}\label{sec:results}
We evaluate accuracy on four medical multiple-choice question datasets in three variants of \name{}: \name{}\textsc{-Mistral1} and \name{}\textsc{-Mistral3}, based on Mistral-7B-Instruct-v0.1 and -v0.3, respectively; and \name{}\textsc{-Llama}, based on Llama 3 Instruct (8B). We report average accuracy and standard deviation across three random seeds. Our results in Tables~\ref{tab:text_graph} and \ref{tab:main} reveal consistent average improvement across datasets compared to baselines. We adopted both LLMs to show that the benefits of our proposal come regardless the underlying model we took as starting point. 

\paragraph{In-prompt graph triples provide useful information}
We investigate whether the inclusion of KG information can positively influence the LLM's answers. To establish a primary baseline, we take Mistral-Instruct-v0.1 and MedQA as a running example. 
For each question, we select a maximum of 10 named entities $s$ and randomly retrieve 2 graph neighbors $o$ for each, resulting in a maximum of 20 graph triples \((s,p,o)\). We include them as part of the prompt, in natural language\footnote{We append the triples at the end of the instruction in JSONL-style, \ie, [\{s1,p1,o1\}, \{s2,p2,o2\}, \ldots].}. Table~\ref{tab:text_graph} shows a degradation in zero-shot accuracy when including triples to the prompt. This can be due to the random selection of the final triples (to fit in the context length), since the semantic information varies among them significantly. This limitation speaks in favor of using KGEs to condense representation of entities to a single embedding. 
Also, as \citet{hager2024evaluation} point out, LLMs' face difficulties in interpreting large amounts of information.
However, fine-tuning the model with triples boosts accuracy, even surpassing BioMistral, a model adapted from Mistral-Instruct-v0.1 through continued pretraining on curated biomedical data. This baseline highlights the value of graph data for the LLM, but it still does not fully leverage the structural and semantic information provided by the KG.

\begin{table*}[t]
    \centering
    \resizebox{0.8\linewidth}{!}{%
        \begin{tabular}{lccccc} 
            \cmidrule(r){1-6}
            ~ & MedQA & PubMedQA & MedMCQA & MMLU-Medical & Avg \\
            \cmidrule(r){1-6}
                        
            \multicolumn{6}{l}{\texttt{Models based on Llama}} \\
            \cmidrule(r){1-6}

            Llama 3 Instruct (8B) &60.3$_{\pm{1.4}}$ &75.8$_{\pm{1.4}}$ &56.4$_{\pm{0.8}}$ &72.8$_{\pm{3.3}}$ &66.3 \\\cmidrule(r){1-6}
            \rowcolor{gray!20} MedAlpaca (7B)$^\dagger$ & 40.1$_{\pm{0.4}}$ & 73.6$_{\pm{0.3}}$ & 37.0$_{\pm{0.3}}$ &55.1$_{\pm{1.1}}$ &51.4 \\
            \rowcolor{gray!20} MEDITRON (7B)$^{\ddagger}$ & 52.0$_{\pm{-}}$ & 74.4$_{\pm{-}}$ & {59.2}$_{\pm{-}}$ &54.2$_{\pm{-}}$ &60.0 \\
            Meerkat-8B$^{\ddagger}$ & \textbf{74.2}$_{\pm{-}}$ & - & \textbf{62.7}$_{\pm{-}}$ & \textbf{75.2}$_{\pm{-}}$ & \textbf{70.7} \\
            RAG-MEDITRON (70B)$^\circ$ &49.6$_{\pm{1.4}}$ &56.4$_{\pm{2.2}}$ &52.7$_{\pm{0.8}}$ &65.4$_{\pm{1.4}}$ & 56.0\\
            RAG-Llama 2 (70B)$^\circ$ &44.9$_{\pm{1.4}}$ &50.4$_{\pm{2.2}}$ &43.1$_{\pm{0.8}}$ &54.6$_{\pm{1.5}}$ & 48.3\\
            \rowcolor{orange!20} \name{}\textsc{-Llama} (8B) &66.0$_{\pm{0.2}}$ &\textbf{78.0}$_{\pm{0.3}}$ &\ul{60.6}$_{\pm{0.3}}$ &\ul{74.9}$_{\pm{0.7}}$ &\ul{69.9} \\
            \cmidrule(r){1-6}
            
            \multicolumn{6}{l}{\texttt{Models based on Mistral 7B}} \\
            \cmidrule(r){1-6}
             
            Mistral-Instruct-v0.1$^\dagger$ & 42.0$_{\pm{0.2}}$ & 73.8$^\triangledown_{\pm{0.4}}$ & 46.1$_{\pm{0.1}}$ &59.1$_{\pm{1.0}}$ &55.3 \\\cmidrule(r){1-6}
            \rowcolor{gray!20} BioMistral$^\dagger$ & 50.6$_{\pm{0.3}}$ &77.5$_{\pm{0.1}}$ & 48.1$_{\pm{0.2}}$ &59.1$_{\pm{1.3}}$ &58.8 \\
            BioMistral DARE$^\dagger$ & 51.1$_{\pm{0.3}}$ &\ul{77.7}$_{\pm{0.1}}$ & 48.7$_{\pm{0.1}}$ &61.9$_{\pm{1.2}}$ &59.9 \\
            Meerkat-7B $^\ddagger$ & \ul{70.3}$_{\pm{-}}$ & - & \ul{60.6}$_{\pm{-}}$ &{70.5}$_{\pm{-}}$ &{67.1} \\
            RAG-Mixtral (8$\times$7B)$^\circ$ &60.0$_{\pm{1.4}}$ &67.6$_{\pm{2.1}}$ &56.4$_{\pm{0.8}}$ &75.9$_{\pm{1.3}}$ &65.0 \\
            \rowcolor{orange!20} \name{}\textsc{-Mistral1} & 54.6$_{\pm{0.2}}$ &74.6$_{\pm{0.6}}$ & 56.4$_{\pm{0.4}}$ &60.3$_{\pm{0.9}}$ &61.5 \\
            \rowcolor{orange!20} \name{}\textsc{-Mistral3} &{60.8}$_{\pm{0.2}}$ &74.4$_{\pm{0.5}}$ & 58.4$_{\pm{0.6}}$ &{68.2}$_{\pm{0.4}}$ &{65.5} \\
            \cmidrule(r){1-6}
        \end{tabular}}
    \caption{Main results on four medical multiple-choice question answering benchmarks. We report accuracy ($\uparrow$) and standard deviation ($\downarrow$), when available, of other 7B and 8B medical open-source models. Avg stands for average across datasets. Llama 3 Instruct and Mistral-Instruct are our base, general purpose models. Models with \hlgray{gray background} are baselines fine-tuned on biomedical data. MEG variants outperformed the respective baselines. 
    $^\dagger$Results reproduced by \citet{labrak-etal-2024-biomistral}. 
    $^\circ$Results from \citet{xiong-etal-2024-benchmarking}.
    $^\triangledown$Reproduced by us with the same data splits used in this work. 
    $^\ddagger$Results from the original papers. 
    }
    \label{tab:main}
\end{table*}

\paragraph{KGE-augmented LLMs show accuracy gains} 
To fully exploit the KG's rich structural information, 
we use node embeddings to guide the LLM's generation. 
This approach also compacts the KG's information into a much shorter input sequence. 
Table~\ref{tab:main} reports accuracy on four  multiple-choice question datasets. 
Across all datasets, \name{}\textsc{-Llama} and both \name{}\textsc{-Mistral} consistently outperform the base LLMs and the fine-tuned baselines (\hlgray{in gray}). 
The only exception is PubMedQA, where BioMistral surpasses both \name{}\textsc{-Mistral}. Its higher accuracy may result from using the artificially labeled training set. Instead, we rather train with the small subset of manually labeled samples to avoid biasing the model towards yes/no answers (see \S~\ref{sec:experiments}). 
The high accuracy on MMLU-Medical indicates that \name{} retains good generalization capabilities. See Appendix~\ref{app:mmlu} for a fine-grained evaluation on MMLU-Medical subjects.
Current SOTA for 7B models, Meerkat-7B, proves the effectiveness of training on chain-of-thought (CoT) in-domain synthetic data. Future work includes exploring CoT instruction tuning in our phase I of training, exploiting information from graph triples instead of relying only on entity labels.
For reference, we also include RAG results from \citet{xiong-etal-2024-benchmarking}. Despite models used by \citet{xiong-etal-2024-benchmarking} are larger than ours and we cannot directly compare to them, accuracy results indicate that MEG is a solid, lightweight alternative.

\subsection{Ablation study}\label{sec:ablation}
In this section, we take \name{}\textsc{-Mistral1} as reference and evaluate the impact of the choice of graph encoder and mapping network architecture on a downstream task (case study on MedQA).

\begin{figure}[t]
    \centering
    \includegraphics[width=\columnwidth]{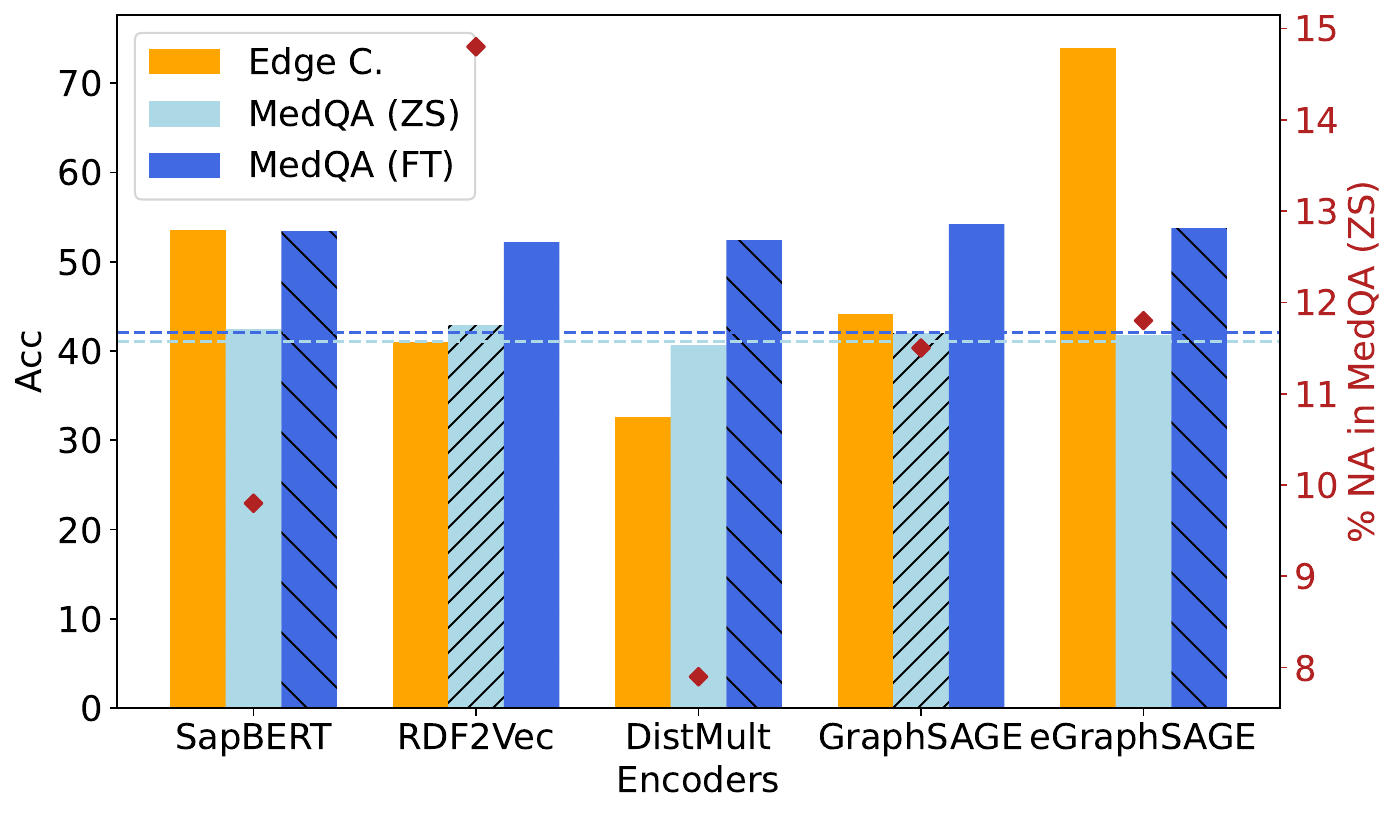}
    \caption{Ablation study on KG encoder choice. Plain bars show edge classification accuracy over UMLS; stripped bars show \name{}\textsc{-Mistral1}'s zero-shot (//) and fine-tuned (\textbackslash\textbackslash) accuracy on MedQA; the dashed line represents accuracy with random embeddings; red dots mark the ratio of not valid answers (NA) in the zero-shot setting.}
    \label{fig:embeds_ablation}
\end{figure}

\paragraph{On the impact of the graph encoder}
We train encoders based on random-walk (RDF2Vec \cite{rdf2vec}), energy (DistMult \cite{distmult}) and message-passing (GraphSAGE \cite{NIPS2017_graphsage} and eGraphSAGE, an edge-type-aware variant inspired by \citet{hu2020pretraining}'s adaptation).
Along with their impact in \name{}'s performance, we include a link classification task as a proxy to evaluate their capabilities. Since these encoders are fundamentally distinct, they capture diverse graph properties, as reflected in classification accuracy in  Figure~\ref{fig:embeds_ablation}, plain (orange) bars. 
eGraphSAGE stands out with a considerably higher score (73.9), as it naturally integrates edge-type information during training. 

However, higher accuracy in a graph-oriented task such as link classification,
does not lead to better performance in a language-oriented downstream task in \name{}. When we integrate these KGEs in \name{}\textsc{-Mistral1} and evaluate zero-shot and fine-tune settings on MedQA (Figure~\ref{fig:embeds_ablation}, stripped bars), eGraphSAGE's substantial advantage in link classification does not carry over to \name{}\textsc{-Mistral1}, as evidenced by the smaller performance gap across encoders (ranging from 52.1 to 54.2). 
This suggests that the role of the KGEs in our setup aligns with our intuition: they guide the answer generation by activating the LLM's semantic region that leads to the correct answer. The difference is more notable in a zero-shot setting, where RDF2Vec produces the highest rate of not valid answers (NA). DistMult's lower NA rate indicates it may better align with the LLM's embedding space. 

\begin{table}[h!]
\centering
\resizebox{\linewidth}{!}{
\begin{tabular}{lccc}
\toprule
~  & \textbf{Parameters (M)} & \textbf{MedQA (ZS)} & \textbf{MedQA (FT)} \\ \midrule
\textbf{MLP 2$\times$512}          & 4.98                    & 41.6                       & 53.5   \\
\textbf{MLP 2$\times$384}          & 3.74                    & 41.5                       & 53.3   \\ 
\textbf{Transf. 4$\times$128}      & 2.18                    & 39.9                       & 53.2   \\ 
\rowcolor{orange!20}\textbf{MLP 4$\times$128}          & 1.22                    & 41.6                       & 54.2   \\ 
\textbf{MLP 3$\times$128 }         & 1.18                    & 42.7                       & 52.7   \\ 
\textbf{MLP 2$\times$128}          & 1.15                    & 38.2                       & 51.3   \\ \bottomrule
\end{tabular}}
\caption{Ablation study on the mapping network done on a separated 1K subset of the train set. Results show accuracy on MedQA for \name{}\textsc{-Mistral1}. ZS: zero-shot; FT: fine-tuned.}
\label{tab:mapping-acc}
\end{table}

\paragraph{On the impact of the mapping network}
To assess the impact of the mapping network architecture
we replace our 4-layer, 128-dimensional MLP (MLP 4$\times$128) with the alternative designs from Table~\ref{tab:mapping-acc}.  
Our final choice, MLP 4$\times$128, outperforms all others in the fine-tuning setting while maintaining a small size. The transformer's lower score highlights the unnecessary overhead of attention layers, as the mapping network’s task of embedding transformation does not benefit from attending the input sequence.

\subsection{Qualitative Analysis}
This section provides insights into the representation spaces before and after the mapping network, comparing them with the LLM's vector space.

\paragraph{Visualizing the embeddings} 
To track the mapped KGEs, we select three UMLS concepts (entities) representing different semantic and specificity levels within the graph's hierarchy, measured by the number of descendants ( \textsc{is\_a} and \textsc{subclass\_of} relations): `Diabetes Mellitus' (CUI: C0011849, a broad disease category), `Headache' (CUI: C0018681, a symptom), and `Atorvastatin Calcium' (CUI: C0286650, a specific pharmacological substance).

\begin{figure*}
    \centering
    \includegraphics[width=\textwidth]{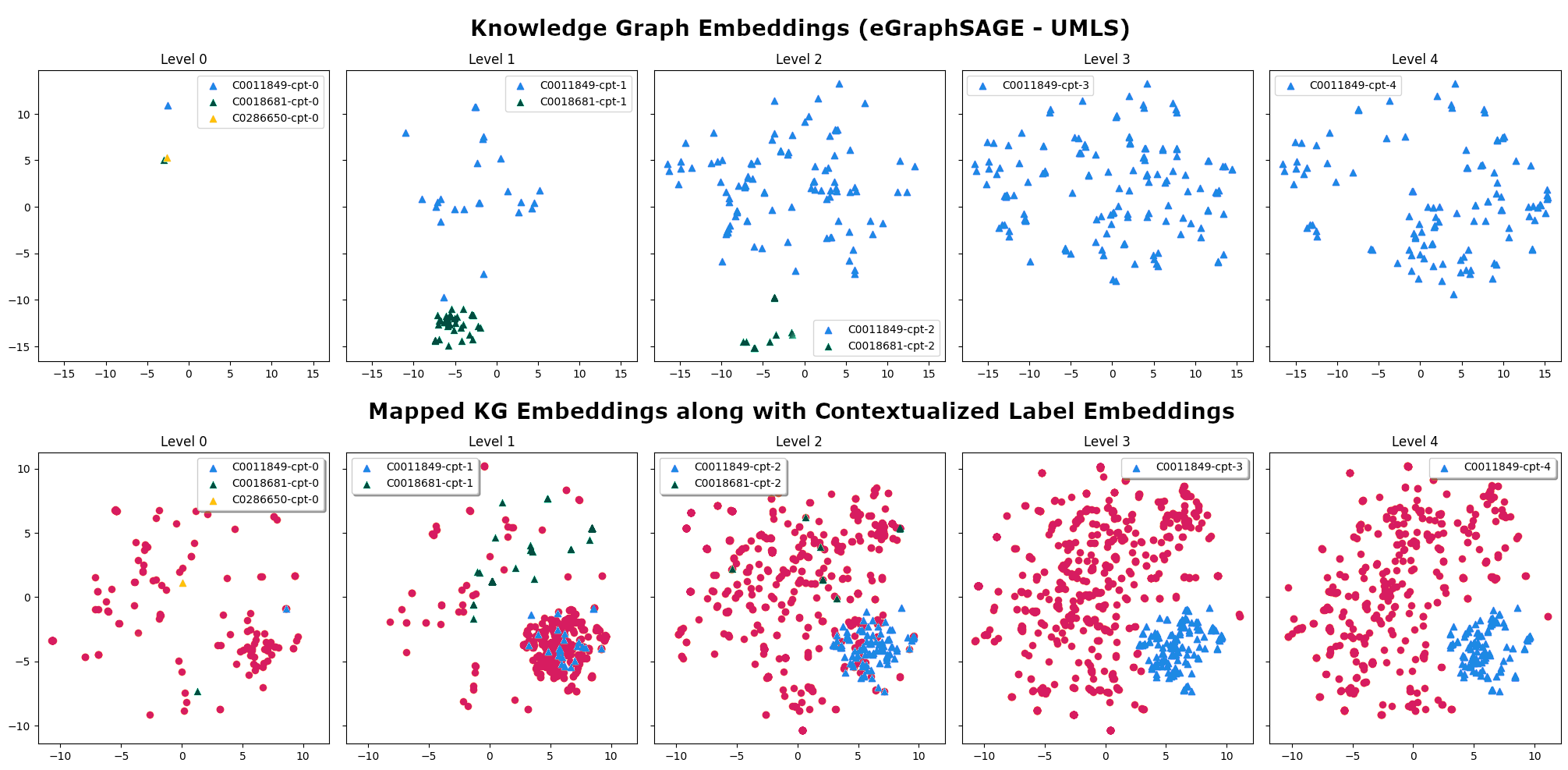}
    \caption{t-SNE visualization of the embeddings: before and after the mapping network. 
    After mapping, the relative KGEs' structure along hierarchy levels is preserved, albeit slightly rotated (\eg, in Level~4, see the diagonal gap which hints the orientation of the blobs) and with reversed sparsity.
    Note the clustering effect over contextualized label embeddings: the mapped KGEs draw them to a specific region.
    }
    \label{fig:visualizing-emb}
\end{figure*} 

Figure~\ref{fig:visualizing-emb} depicts t-SNE\footnote{We have also witnessed the same overall behavior using UMAP and MDS to visualize the embeddings.} plots of the concepts (Level 0) and their hierarchies (Level~1 to ~4) in the KGE space (top row) and the mapped KGE along with the concept's contextualized embeddings in the LLM space (bottom row). Such contextualized embeddings are the LLM's token embeddings generated for the labels of the concepts when verbalizing their KG's triples, \ie, they represent the concepts in all the contexts given by the KG. 

Examining the distribution of concepts in the upper and lower rows, we observe two effects of the mapping. 
First, the relative structure of KGEs is preserved after mapping to the LLM's space, albeit slightly rotated (see Level~1 and Level~4) and with reversed sparsity
(high-density groups become less dense, while low-density groups tighten; yet the global between-group structure is the same)\footnote{Given that KGE's and LLM's embeddings represent two spaces with different dimensionalities, we applied t-SNE separately to each; thus, a difference in scale is expected.}. 
Second, the
clustering effect of the contextualized label embeddings (see Level~0): while the labels (circles in the bottom row) are scattered, the mapped embeddings cover such spaces. 
As we descend in the graph's hierarchy, 
the contextualized labels become increasingly scattered, yet the mapped KGEs, while remaining within the same region, stay more cohesive.
Note that these contextualized labels are 
distinct vectors for the same label across different contexts. Indeed, their sparsity suggests that they do not completely capture the conceptual dimension of the entity they represent.
The mapping seems to activate the LLM's embedding region associated with each concept, providing semantic information the model uses to condition its answer generation.

\paragraph{Probing \name{} for acquired knowledge}
We informally probe \name{} to estimate its acquired medical knowledge. We handcraft a set of open-ended questions to query Mistral-Instruct before and after the phase I training on UMLS (\S~\ref{sec:training}), and \name{}\textsc{-Mistral1}. As an example, we show an excerpt (irrelevant information to the question is omitted for brevity) of models' answers to the question \textit{`What are the active and inactive compounds present in Atorvastatin?'}\footnote{This question targets Atorvastatin (C0286650 in UMLS).} and \hlyellow{highlight} the key differences. \name{} uses the KGE for Atorvastatin and provides a more knowledgeable response.

\noindent
\textbf{Mistral-Instruct (baseline):} \ldots The active compound in Atorvastatin is \hlyellow{Atorvastatin itself}, while the inactive compounds are \hlyellow{Atorvastatin calcium salt} and \hlyellow{Atorvastatin magnesium salt}. 

\noindent
\textbf{Mistral-Instruct after phase I:} \ldots The active compound present in Atorvastatin is \hlyellow{Atorvastatin calcium}. 
\ldots is responsible for the drug's therapeutic effect. 
\hlyellow{There are no inactive compounds present in Atorvastatin}.

\noindent
\textbf{\name{}\textsc{-Mistral1}:} \ldots the active compound \hlyellow{Atorvastatin calcium}.The inactive compounds present in Atorvastatin include \hlyellow{excipients such as microcrystalline cellulose, croscarmellose sodium, magnesium stearate, and sodium lauryl sulfate} 
\ldots 

\section{Discussion and Future Research Lines}
Our approach is efficient not only because it trains a small fraction of the LLM's parameters but also because it seamlessly handles out-of-vocabulary terms. 
We present and efficient approach to domain specialization of models, as it updates only a small fraction of the LLM's parameters while seamlessly handling out-of-vocabulary terms.
A new term's KGE can be initialized, for instance, by averaging its one-hop neighbors in the KG, making \name{} both lightweight and adaptable to new vocabulary. However, the efficacy of this method should be evaluated in future work.

\citet{garikipati2024openmedlm} demonstrate that prompt engineering can outperform fine-tuning in medical QA for open-source LLMs. However, our focus was to investigate the viability of integrating knowledge from KG embeddings into LLMs rather than optimizing for peak downstream performance. Our experiments show that supervised fine-tuning of KGE-augmented LLMs yields more accurate answers than other specialized baselines. 
Chain-of-thought tuning, as shown by \citet{kim2024smalllanguagemodelslearn}, is another promising step forward to improve \name{}'s accuracy.
\name{} improves response generation by injecting KGEs in a single generation step. 
This suggests that \name{} could also benefit from chain-of-thought tuning, as each of the reasoning steps would increase precision of the model's response.

Besides, the sensitivity of LLMs to the information order in multiple-choice questions, also known as positional bias, is well-documented \cite{pezeshkpour2023largelanguagemodelssensitivity,zheng2024large}. More specific to biomedicine, \citet{LIEVIN2024100943} and \citet{hager2024evaluation} show how variations in sequence can significantly impact diagnostic accuracy of medical-aligned language models. However, the robustness to changes in information order remain understudied in most medical model evaluations. Recent studies \cite{wang-etal-2024-answer-c,wang2024look} find that greedy decoding combined with text answer evaluation gives more consistent answers compared to first-token evaluation, particularly for instruction-tuned LLMs. Thus, in an attempt to alleviate this issue, we inspect the text answer generated by the model instead of ranking the candidate answers by the log probability of its first token prediction.
However, whether KGEs further help the LLMs in mitigating positional biases---as well as other \cite{lyell2017automation,moor2023foundation,ness2024medfuzzexploringrobustnesslarge} biases---needs to be explored in future work.


\section{Conclusion}
We introduce \name{}, a novel medical knowledge-augmented LLM based on KGEs for question answering tasks. To the best of our knowledge, we are the first to inject pretrained KGEs into an LLM via a lightweight mapping network. 
The LLM interprets the graph information encoded in KGEs, which further guides the response generation.
We present a comprehensive evaluation on four medical multiple-choice question benchmarks, revealing that LLMs can highly benefit from the factual information encoded in KG embeddings. Our results suggest that integrating KGEs with LLMs offers a promising path towards specialized language models. 



\bibliography{tacl2021}
\bibliographystyle{acl_natbib}


\appendix
\clearpage

\section{Training Details}
\label{app:training_details}
\subsection{Phase I: Embedding Transfer Learning}
Since the input prompt has a fixed size (see Figure~\ref{fig:mapping}), we use a reduced sequence length of 124 to optimize computational efficiency. We train for one epoch with gradient accumulation over 8 steps to achieve an effective batch size of 128. We employ a cosine learning rate scheduler with learning rate of $1e-5$, warmup ratio of 3\% and no weight decay. We use mixed precision (bfloat16) and FlashAttention2 \cite{dao2023flashattention2fasterattentionbetter} to optimize memory usage and speed up computations on the LLM. 

\subsection{Fine-tuning on Downstream Tasks}
Fine-tuning is done for 3 epochs, with a sequence length of 400 (500 for PubMedQA), learning rate of $1e-4$ and effective batch size of 32. The remaining hyperparameters have the same value as in Phase I. We did not optimize hyperparameters for Llama-3-Instruct.  Total number of trainable parameters: 83M, in both \name{}\textsc{-Mistral} and \name{}\textsc{-Llama}.

\section{Extended Results}
\label{app:mmlu}
\begin{table*}[h!]
    \centering
    \resizebox{0.8\linewidth}{!}{%
        \begin{tabular}{lcccccc} 
            \cmidrule(l){1-7}
            ~ & Clinical K. & Genetics & Anatomy & P. Medicine & C. Biology & C. Medicine \\
            \cmidrule(l){1-7}
            
            \multicolumn{6}{l}{\texttt{Models based on Llama}} \\
            \cmidrule(l){1-7}

            LLama 3 Instruct (8B) &73.6$_{\pm{2.7}}$ &85.0$_{\pm{3.6}}$ &68.9$_{\pm{4.0}}$ &69.1$_{\pm{2.8}}$ &78.5$_{\pm{3.4}}$ &61.8$_{\pm{3.7}}$\\\cmidrule(r){1-7}
            \rowcolor{gray!20}MedAlpaca (7B)$^\dagger$ & 53.1$_{\pm{0.9}}$ & 58.0$_{\pm{2.2}}$ & 54.1$_{\pm{1.6}}$ & 58.8$_{\pm{0.3}}$ & 58.1$_{\pm{1.3}}$ & 48.6$_{\pm{0.5}}$ \\
            \rowcolor{gray!20}MEDITRON-7B$^{*\ddagger}$ & 57.2$_{\pm{-}}$ & 64.6$_{\pm{-}}$ & 49.3$_{\pm{-}}$ & 55.4$_{\pm{-}}$ & 53.8$_{\pm{-}}$ & 44.8$_{\pm{-}}$ \\
            Meerkat-8B$^{*\ddagger}$ & 74.3$_{\pm{-}}$ & 76.7$_{\pm{-}}$ & 74.8$_{\pm{-}}$ & 75.3$_{\pm{-}}$ & 76.1$_{\pm{-}}$ & 74.3$_{\pm{-}}$ \\
            \rowcolor{orange!20}\name{}\textsc{-Llama} (8B) & 72.3$_{\pm{0.5}}$ & 83.0$_{\pm{1.5}}$ & 64.5$_{\pm{0.7}}$ & 79.4$_{\pm{0.3}}$ & 80.6$_{\pm{0.4}}$ & 69.4$_{\pm{0.9}}$ \\
            \cmidrule(l){1-7}
            
            \multicolumn{6}{l}{\texttt{Models based on Mistral 7B}} \\
            \cmidrule(l){1-7}
             
            Mistral-Instruct-v0.1$^\dagger$ & 62.9$_{\pm{0.2}}$ & 57.0$_{\pm{0.8}}$ & 55.6$_{\pm{1.0}}$ & 59.4$_{\pm{0.6}}$ & 62.5$_{\pm{1.0}}$ & 57.2$_{\pm{2.1}}$ \\
            \rowcolor{gray!20}BioMistral$^\dagger$ & 59.9$_{\pm{1.2}}$ & 64.0$_{\pm{1.6}}$ & 56.5$_{\pm{1.8}}$ & 60.4$_{\pm{0.5}}$ & 59.0$_{\pm{1.5}}$ & 54.7$_{\pm{1.0}}$ \\
            BioMistral DARE$^\dagger$ & 62.3$_{\pm{1.3}}$ & 67.0$_{\pm{1.6}}$ & 55.8$_{\pm{0.9}}$ & 61.4$_{\pm{0.3}}$ & 66.9$_{\pm{2.3}}$ & 58.0$_{\pm{0.5}}$ \\
            Meerkat-7B $^\ddagger$ & 71.6$_{\pm{-}}$ & 74.8$_{\pm{-}}$ & 63.2$_{\pm{-}}$ & 77.3$_{\pm{-}}$ & 70.8$_{\pm{-}}$ & 65.2$_{\pm{-}}$ \\
            \cmidrule(l){1-7}
            
            \rowcolor{orange!20}\name{}\textsc{-Mistral1} & 58.1$_{\pm{0.8}}$ & 68.7$_{\pm{0.2}}$ & 54.4$_{\pm{0.5}}$ & 62.9$_{\pm{0.9}}$ & 61.1$_{\pm{2.2}}$ & 56.6$_{\pm{1.0}}$ \\
            \rowcolor{orange!20}\name{}\textsc{-Mistral3} & 64.9$_{\pm{0.2}}$ & 69.6$_{\pm{0.8}}$ & 63.0$_{\pm{1.0}}$ & 72.8$_{\pm{0.4}}$ & 73.6$_{\pm{0.0}}$ & 65.2$_{\pm{0.2}}$ \\
            \cmidrule(l){1-7}
        \end{tabular}}
    \caption{Performance on six individual MMLU-Medical categories. We report accuracy ($\uparrow$) and standard deviation ($\downarrow$), when available, of other 7B and 8B medical open-source models. Models with \hlgray{gray background} are baselines fine-tuned on biomedical data. }
\end{table*}


\end{document}